# Chapter 1.4

# Design Perspectives of Multitask Deep Learning Models and Applications

**Yeshwant Singh[1*], Anupam Biswas[2], Angshuman Bora[3], Debashish Malakar[4], Subham Chakraborty[5], Suman Bera[6]**
Department of Computer Science and Engineering,
National Institute of Technology, Silchar, Assam, India, 788010
Email: *[1]yeshwant_rs@cse.nits.ac.in, [2]anupam@cse.nits.ac.in, [3]angshu.btf@gmail.com, [4]debasmkr555@gmail.com, [5]subham.chakra07@gmail.com, [6]sumanreborn18@gmail.com

**Abstract-** In recent years, multi-task learning has turned out to be of great success in various applications. Though single model training has promised great results throughout these years, it ignores valuable information that might help us estimate a metric better. Under learning-related tasks, multi-task learning has been able to generalize the models even better. We try to enhance the feature mapping of the multi-tasking models by sharing features among related tasks and inductive transfer learning. Also, our interest is in learning the task relationships among various tasks for acquiring better benefits from multi-task learning. In this chapter, our objective is to visualize the existing multi-tasking models, compare their performances, the methods used to evaluate the performance of the multi-tasking models, discuss the problems faced during the design and implementation of these models in various domains, and the advantages and milestones achieved by them.

## 1.4.1. Introduction

Machine learning optimizes a specific measurement metric irrespective of its physical value on a particular standard or a business key performance index. Single-task or multiple single-task models are trained to perform well on new data. The parameters of the models are calibrated to increase the performance metrics until they cannot improve further. We may accomplish satisfactory results by pinpointing a single task, but we may overlook data that could generalize models well in the process. The transformation performed by multi-task neural network hidden layers on data originates a data representation by overlapping multi-task signals. The model can fully summarize the initial task by distributing information among related tasks. This method is termed Multi-Task Learning (MTL). The difference between single-task learning compared to MTL is given in **Fig. 1.4.1**.

The motivation behind MTL comes from various directions. For example, Humans learn new things in various fields during their life; we regularly employ the information we have accumulated by learning from similar past experiences or tasks. A newborn starts learning how to recognize different people's faces at an early age. The knowledge/information gained is invariant and applicable to other world objects. Moreover, from an instructional point of view, we regularly learn new tasks based on the collective experience gained over other tasks. This makes us perform complex tasks.

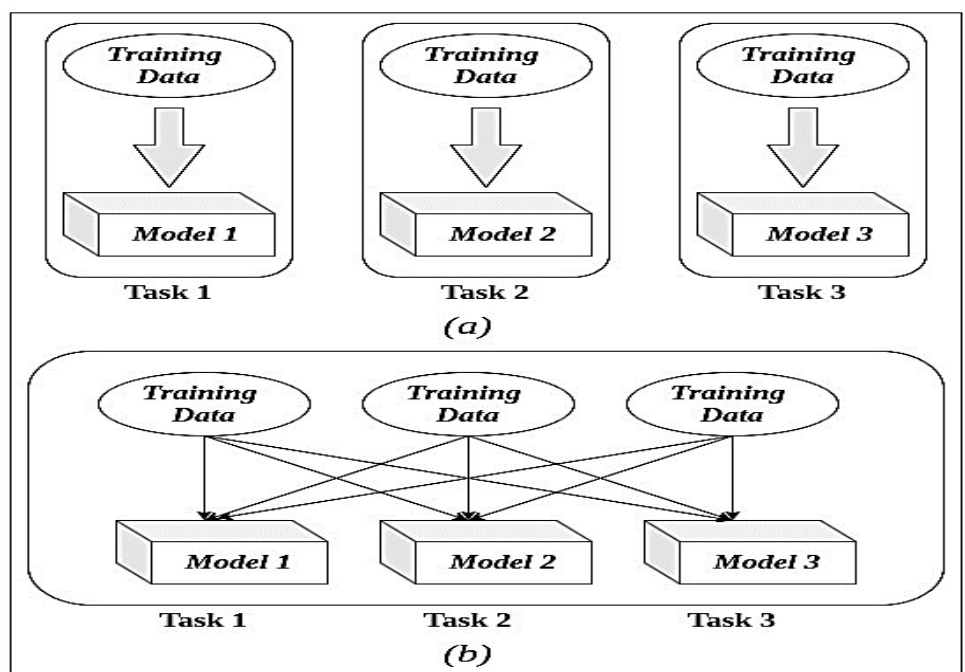

**Fig. 1.4.1**: Single-Task Learning versus Multi-Task Learning Diagram

Some of the practical applications of MTL are web search, spam filtering, etc., [1]. In web search, MTL with boosted decision trees is incredibly supportive as data sets from multiple countries differ significantly in size due to the expense of editorial judgments. It has been demonstrated that learning different overlapping tasks can mutually commence considerable improvements in accuracy with quality.

Spam filtering can be treated as particular but related classification tasks over a set of users. In other words, an email considered spam to one user might not appear spam to someone else. The definition of spam emails varies from person to person, yet spam classification has related tasks. Solving every individual user spam classification jointly with the help of MTL can improve all users' performance.

In this chapter, we attempt to summarize MTL from a design perspective. We begin by defining deep learning in general in Section 1.4.2 and its relation to MTL. Then we go into some popular classes of neural networks in Section 1.4.3 to give a base foundation of underlying ideas. Section 1.4.4 gives an overview of the designs and implementation of various methods in MTL. Section 1.4.5 presents a few applications of MTL in some domains, followed by evaluation methods used to measure the performances of these models in Section 1.4.6. Finally, we conclude this chapter in Section 1.4.7 with challenges and future directions of MTL.

### 1.4.2. Deep Learning

Deep Learning (DL) is a sub-domain of Machine Learning (ML). It is entirely made up of artificial neural networks aided representation learning. Here, learning is classified into three types: supervised, semi-supervised, and unsupervised [2]. The term "deep" originates from using multiple layers, including hidden layers in the Neural Networks (NN). MTL falls in the sub-domain of deep learning.

In the human brain, there are over 100 billion interconnected Neurons. These are the basic working units of the brain. Every Neuron is inter-linked through thousands of Synapses to neighboring Neurons in a giant web-like structure. In DL, Neurons are represented by formulating an artificial neural net consisting of nodes (Neurons). A Neuron takes an input, performs an operation, and gives an output. The operation performed over the input is a linear transformation, making the operation highly parallelable and efficient to compute. Non-linearity is added to the computation of a Neuron in terms of the activation function. This non-linearity allows the neural network to learn non-linear patterns from the data. The diagram of a neuron in the brain versus an artificial neural net is shown in **Fig. 1.4.2**.

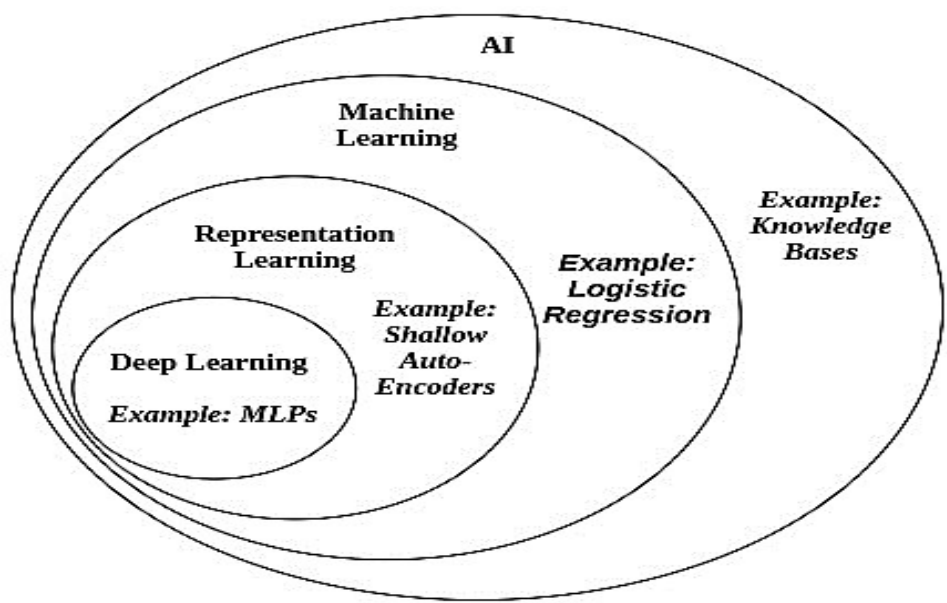

**Fig. 1.4.2:** AI Landscape and sub-emerging fields

Some real-life applications of DL are Automatic Text Generation, here the neural network model first learns from the corpus of text. It learns to spell, form sentences, and the style of writing. After that, the model can generate new texts. Another field where DL is achieving significant research is Image Recognition; this field deals with how computers can understand and interpret various images or videos. DL-trained models fitted in vehicles can now take input from 360° Camera [3] to make automatic driving possible.

### 1.4.2.1 Feed-Forward Neural Network

Generally, the fully connected layer is the final layer in neural network architectures. A feed-forward neural network (FNN) is also called a Multi-layer perceptron (MLP). The input to this layer is the output of the previous layer (pooling or convolution), then it is flattened into a one-dimensional input vector. An activation function, for example, the Soft-max activation function, is applied to the inputs used as input of the first layer of the fully connected layer. The last layer of this part classifies the images based on the probability of an image that belongs to a particular class. Weights are initialized to the network and multiplied by each input element to determine the probabilities.

### 1.4.2.2. Convolution Neural Network

Convolution Neural Network (CNN) is a specialized type of Deep Neural Network (DNN) model primarily used for working with images (1D, 2D, 3D) to extract the essential features present in the image data using trainable filters. It takes images as inputs to modify the learnable parameters to properly distinguish the salient features and capture their spatial and temporal dependencies. The computation and processing of CNN are much less and efficient than most traditional image classification algorithms.

### 1.4.2.2.1. Convolution Layer

It is the process of filtering the image with weighted averaging known as the filter kernel. The filter kernel, also known as the convolution matrix, is a small matrix of a given size applied to the image to extract out the desired features. After convolution of the image with the filter kernel, the output is called the feature map. Let I be the image and F be the kernel, and IF, as denoted in the equation (Eq. (1)), be the convolution result of applying F kernel over I image.

$$IF(m, n) = \sum_{i,j} I(m - i, n - j) . F(i, j) \qquad (1)$$

**1.4.2.2.2. Pooling Layer**

CNNs use pooling layers in addition to the convolution layers. This layer is applied in between two convolution layers. This layer down-samples patches of the feature map into a summarized value based on the type of pooling applied over the whole feature map. The size of the pooling filter is less than that of feature maps. If its size is 2x2, it wills down-sample the dimension of the feature map to one-quarter of its size. This helps in reducing the dimension of the image data without losing its salient features and thus reduces the computational time. The types of pooling used in CNN are:

*a) Average Pooling:* This method considers the average value for every patch in the feature map. Then, the down-sampling is done following the average value.

*b) Max Pooling:* Here the maximum of each patch in the feature map is calculated. Then, the patch is down-sampled with the maximum value calculated.

**1.4.2.3. Recurrent Neural Network**

Recurrent Neural Networks (RNNs) are a significant category of Neural Networks. It came from a feed-forward neural network; it can utilize its hidden (internal) to deal with sequences inputs of various lengths. The inner hidden layer acts as a memory bank that helps remember the information already calculated. The "recurrent neural network" generally refers to two broad classes of neural networks with nearly equivalent overall structures: finite impulses and infinite impulses.

In traditional Neural Networks, every input sample and corresponding output are considered unbiased of each other. However, some cases require anticipating the following words of a sentence or any sequence. The preceding phrases are needed, and there is a need to recollect the previous phrases solely for this reason. Consequently, RNNs came into relevance; they did solve the problem with their hidden layers. The hidden state remembers a few accounts approximately a sequence.

The significant applications of RNNs are handwriting recognition [4] and speech recognition [5]. Long short-term memory (LSTM) is a prevalent RNN model, primarily due to its feedback connections [6]. It is used in sequence prediction problems.

### 1.4.3. Multi-Task Deep Learning Models

MTL is a sub-domain of machine learning that focuses on processing multiple (more than a single) tasks simultaneously by considering the relevance among the various tasks. The deep learning model focuses only on one output attribute in single-task learning by defining the problem as a regression, classification, or prediction problem. On the other hand, in MTL, the model focuses on multiple output attributes. Various problems are defined either as the same or a mix of regression, classification, and prediction problems. i.e., multiple relevant classifications or prediction problems can be solved with a single multi-task model considering multiple output attributes for each problem. This section discusses different kinds of multi-task deep learning models from the types of problems they solve.

#### 1.4.3.1. Classification Models

Solving different problems as classification problem is quite common in deep learning. However, unlike single-task learning, the model is designed for multiple output attributes as several classification tasks in multi-task deep learning.

#### 1.4.3.1.1. Multi-attribute recognition models using joint learning of features

The objects present in an image have multiple attributes. Let an image 'I' consist of an object with A = {a1, a2, a3, a4, ..., an} number of attributes. The task is to design a model which determines these attributes present in the image. Some of the attributes are correlated with each other of different objects of the same class. The features must be shared with all other sub-tasks to advance the model's learning and accuracy. Therefore, correlated attributes are advantageous for the multi-task deep learning models. A straightforward approach will be to design a model using a multi-class classification neural network. However, variations in the attribute classes hinder the model's performance. The main idea is to learn shared feature representations and a shared model simultaneously, as shown in **Fig. 1.4.3**.

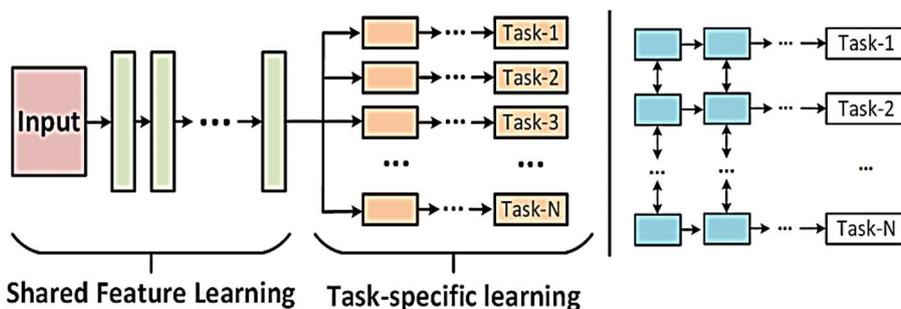

**Fig. 1.4.1:** Hard parameter (left) and soft parameter (right) sharing MTL using joint feature learning

Li et al. [7] have developed a joint learning-based multi-task DL model to overcome this issue. The use of MTL is an efficient approach to tackle this problem domain. The objects present in an image have multiple attributes, and some of the attributes correlate with each other of different objects of the same class. The features must be shared with all other sub-tasks to advance the model's learning and accuracy. This can be done using feature joint learning in MTL [7]. Hard and soft parameter sharing are the two MTL models that can train the model. The designing of the model consists of a series of steps, including image data pre-processing, feature extraction, using a classifier, and dividing into 'n' number of sub-task classifiers to create a pipeline using MTL and thresholding the classified outputs to identify all the attributes at once [8]. **Fig. 1.4.4** shows the flowchart of R-CNN for landmark locations, face detection, gender detection and pose estimation.

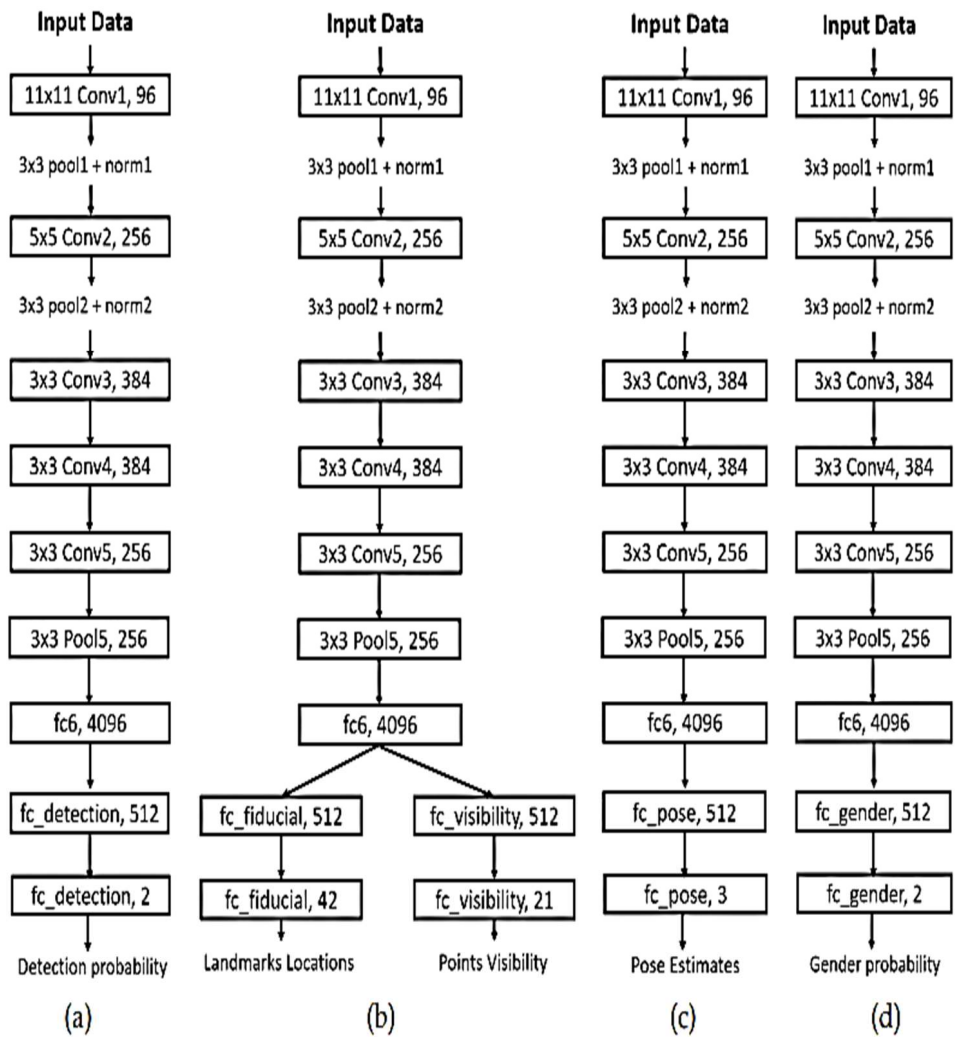

**Fig. 1.4.4**: R-CNN for landmark locations, face detection, gender detection and pose estimation.

### 1.4.3.1.2. Multi-task Facial attributes classification model using feature fusion

The Convolution Neural Network architecture comprises several layers, which involve many activation maps. This increases the dimensionality of the hyper features [9]. The hyper features need to be linked to retrieve the features similar to the multiple tasks in an efficient manner. One of the basic approaches to address this kind of problem is using feature fusion which combines feature vectors to get a single feature vector. As CNN can evaluate complex functions, it can perform a fusion of the hyper-parameters. Region-based CNNs (R-CNNs) are the network architectures used for face detection, gender detection, and pose estimation [10], as shown in **Fig. 1.4.4**. The R-CNN architectures for the given classification problems are shown in **Fig. 1.4.5**.

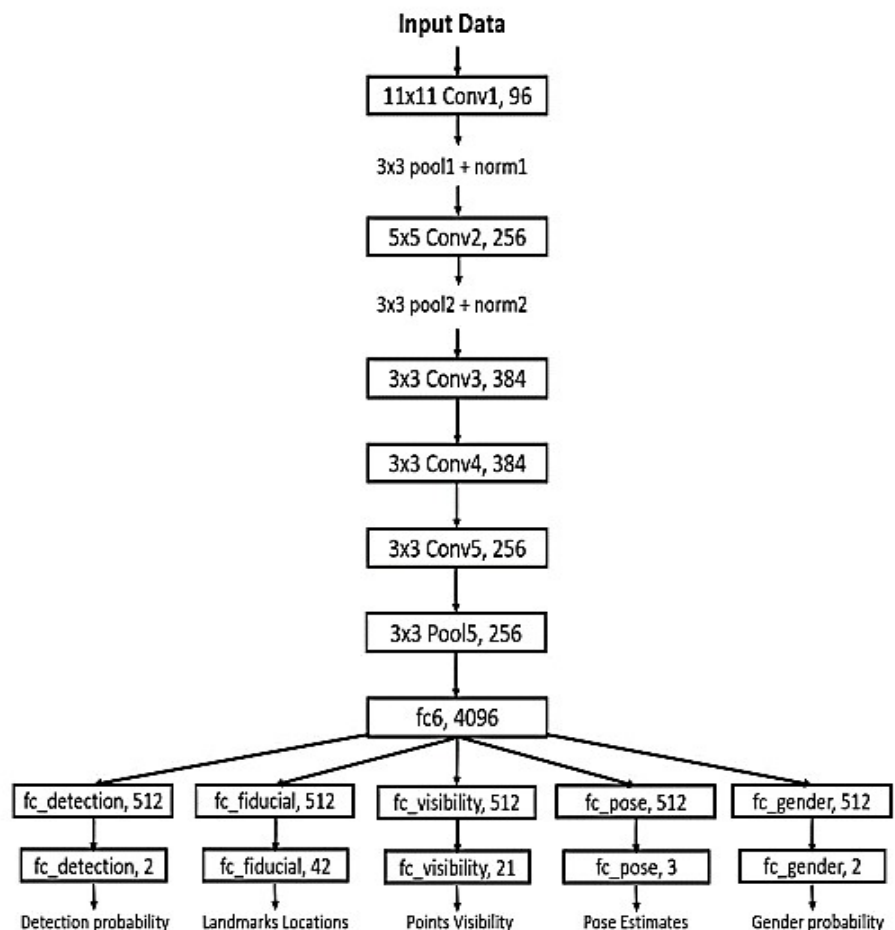

**Fig. 1.4.5:** MTL architecture by combination of several R-CNN

### 1.4.3.2. Prediction Models

### 1.4.3.2.1. Multi-tasking on time-series data

The model is built to predict clinical time-series data, including mortality risk, length-of-stay prediction, decomposition, and phenotype classification [11]. Then by using multi-task learning, all these tasks are trained. Each of its tasks is different from the other in terms of output and structure. The use of LSTM neural network architecture using multi-tasking improves prediction than the linear regression models. The time-series data is being re-sampled to regularly spaced intervals. Channel-wise LSTM, which is a modified version of the LSTM baseline, is used with MTL as the regularizer over each task, as shown in **Fig. 1.4.6**. For prediction purposes, L2 regularization is being used. The formula for L2 regularization is given below in the equation (Eq. (2)):

$$L(x,y) = \sum_{i=1}^{n}(y_i - h_\theta(x_i))^2 + [\lambda \sum_{i=1}^{n} \theta_i^2] \qquad (2)$$

The term in the square bracket is the square magnitude of coefficients added to the loss function. The final loss in the network is the summation of all the task-specific losses.

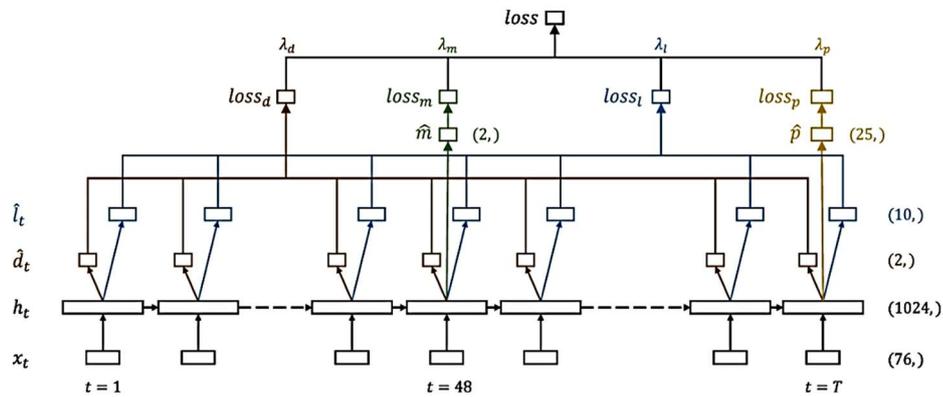

**Fig. 1.4.6:** LSTM multi-task architecture

**1.4.3.2.2. Multi-step forecasting on multivariate time-series using split layers**

In multivariate time series, the tasks are divided mainly into primary and additional tasks. The primary tasks are needed to be formulated for a target series. The additional tasks are needed as the hidden representation improves prediction accuracy. The training of this model follows a split-network architecture which is expanded into the feature that can be shared and cannot be shared among the primary and supplementary tasks in this multivariate time series. Using split architecture allows extending the number of task-specific representations and shared representations [12]. Split layers properly stabilize the hidden and task-specific representations in the primary and supplementary series. **Fig. 1.4.7** shows the representation of the architecture of the split layers. The use of different colors indicates the different task-specific layers. The supplementary targets are weighted less compared to the primary targets so that the model focuses on handling the errors in primary targets more than the supplementary targets.

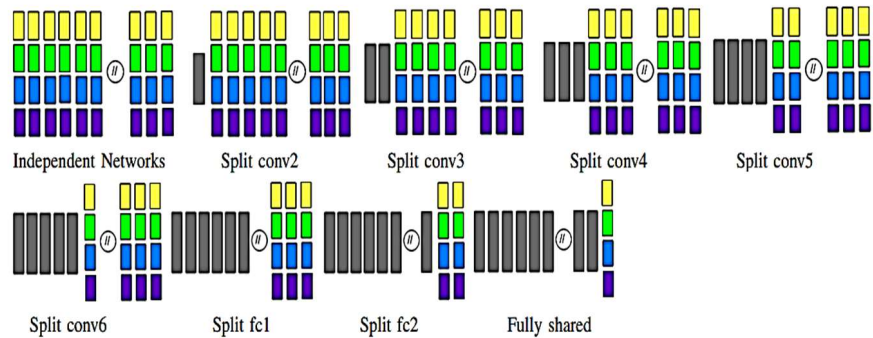

**Fig. 1.4.7**: Network architecture of the split layers

### 1.4.3.3. Mixed Models

Basic statistics depend mainly on normally distributed data, but these basic statistical methods fail when the data sets fall outside the range. Elementary statistical methods tend to assess only the exact effects shown by the predictor variables, while the problem domain often involves random effects [13]. The most common random effects are visible in the parts of the experimental studies that are replicated spatially. Diversification among individuals is also observed due to random effects. Researchers often ignore the possibility of random effects or use them as fixed factors, which might violate statistical assumptions and reduce the scope of inference. There is an excellent necessity for mixed models in statistical analysis to incorporate random effects in the model. Mixed models are typically used when statistical observations or samples are clustered, especially in ecology and biomedical sciences. Batch effects in biomedical sciences lead to non-independence of statistical observations. We can use a naive linear t to x the linear regression coefficients in the fixed-effects model. However, random effects, i.e., slope and intercept, are no longer fixed but vary around their mean values in individual levels when it comes to the coefficients.

Let us consider a sleep deprivation study where the sleeping schedule of 18 individuals was restricted, and their reactions were observed. Their reaction, days, and subjects were taken for ten days. An ordinary least square linear regression was fitted to check the change of response of individuals to sleep deprivation. The reaction versus day's graph, which was made by fitting ordinary least square linear regression, has an increasing trend, but much variation has been observed between days and individuals. It is assumed that all the observations were uncorrelated and hence normally distributed. However, the data points within individuals are not independent instead clustered. When a Linear Mixed Model was fitted with random slopes and intercepts for each individual, two statistics were observed, one for fixed effects and another for random effects. The random effect statistics fixed the error observed between the non-independent samples.

Simpson et al. [14] extended a broad approach to mixed modeling to study system-level properties of the brain across multiple tasks. It allows estimating population network differences between tasks, their relations to outcomes, and evaluating individual variability in network differences. A two-part mixed model is implemented if there is a connection for estimating the strength and probability of the connection [15]. The whole matrix of brain connectivity for every participant comprises the overall model. The endogenous covariates are abstract variables obtained from the network to describe the global topology, whereas the exogenous covariates predict the physiologically related phenotypic traits. The above described statistical framework provides decent results in group and individual effects. When the formulation of multivariate statistics is provided, the framework accounts for interdependence among the edges present in a network.

The correlation between the covariates and the corresponding tasks of network connectivity can be explained using this statistical framework. It also analyses network connectivity across groups and tasks and forecasts network connectivity based on nodal network features and task status. This model also provides a more inclusive understanding of topological variability and a medium of measuring goodness-of-t. It provides information related to the changes that occur in the brain during task changes and transitions. Hence mixed effect models provide a better insight into task analysis in multi-tasking.

### 1.4.4. Design and Implementation
### 1.4.4.1. Multi-Task Learning Methods used for Deep Learning

Soft and hard parameter sharing are the systematic methods employed in MTL. The parameters are shared from the hidden layers.

### 1.4.4.1.1. Hard Parameter Sharing

It is one of the frequently used methods to perform MTL in Deep Learning [16]. The particular method is applied to each task by sharing hidden layers and maintaining many output layers specifically for each task.

Over-fitting of the model decreases to a large extent by Hard parameter sharing. The author [17] has shown the danger of over-fitting is of an order N (total number of tasks). We can also see this intuitively in a model that learns more than one task simultaneously, i.e., the

MTL model also has to find a representation that will capture each task. This results in overall less risk concerning over-fitting the actual task. The general architecture of hard parameter sharing is shown in **Fig. 1.4.8.**

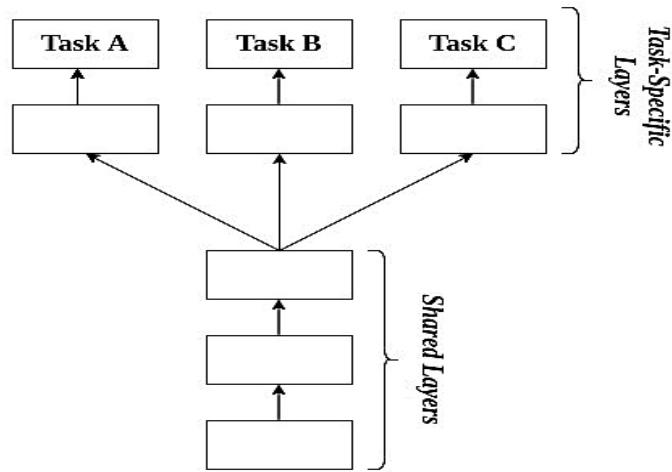

**Fig. 1.4.8**: Hard Parameter Sharing

**1.4.4.1.2. Soft Parameter Sharing**

While sharing Soft parameters, every task has a deep learning model. Here, the parameters are assumed to be similar in the multi-task model; thereby, the parameters are regularized. Different normalization techniques are used for regularization. Doung et al. [18] used the L2 norm, while Yang and Hospedales [19] used the trace norm. Regularization techniques in MTL developed for other deep learning models have greatly encouraged the constraints of sharing soft parameters, as shown in **Fig. 1.4.9**.

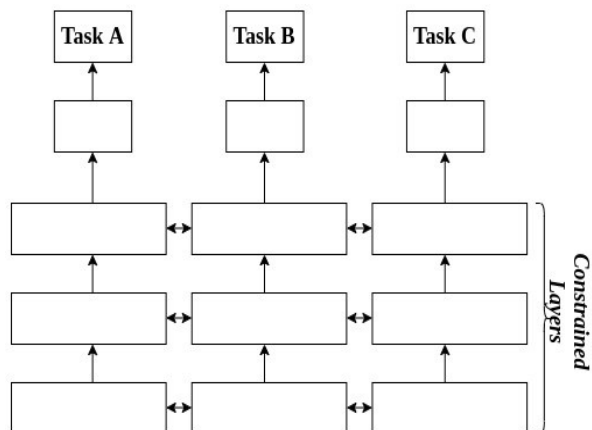

**Fig. 1.4.9**: Soft Parameter Sharing

### 1.4.4.2. Various Design of Multi-task Learning

### 1.4.4.2.1. Deep Relationship Networks

Computer vision in MTL is commonly approached by sharing the convolution neural network layers. After which, the model learns the task-specific work using fully connected layers. The approach by Long et al. [20] proposed a methodology to enhance specific models by using Deep Relationship Networks, along with the shared layers and specific layers for each task of MTL, as shown in **Fig. 1.4.10**. A matrix was placed before the fully connected layers that allowed the deep learning model to learn various relationships among the tasks. This method is comparable to some Bayesian models. However, there is a drawback here. This proposed methodology still depends on a predefined structure to share different parameters.

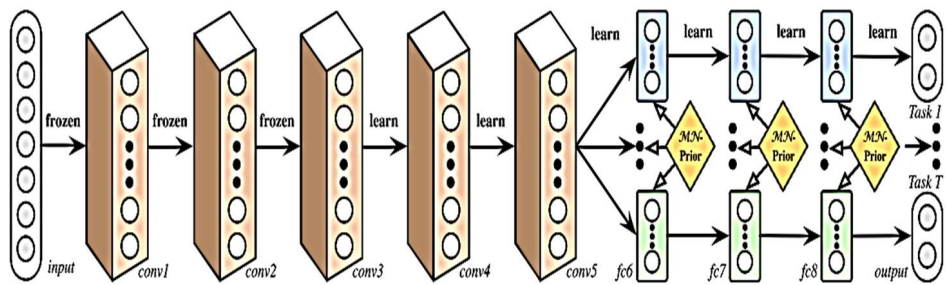

**Fig. 1.4.10**: Deep Relationship Network [20]

### 1.4.4.2.2. Fully Adaptive Feature Sharing

Another intriguing technique described by Lu et al. [21] is an upside-down strategy that uses a small neural network at the beginning. Then eager broadens itself during training using a fundamental notion that supports comparable grouping tasks. The method for widening is to create branches dynamically, as shown in **Fig. 1.4.11**. However, this greedy approach method might fail to and a globally optimal model. This approach does not permit the model for learning the intricate interactions within different tasks by assigning each branch precisely one task.

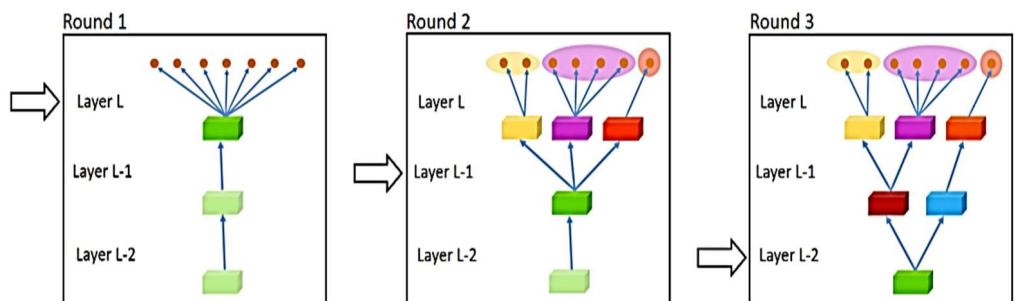

**Fig. 1.4.11**: Fully Adaptive Feature Sharing [21]

**1.4.4.2.3. Cross-stitch Networks**

Misra et al. [22] proposed a model with two different architectures, almost the same as in soft parameter sharing. They proposed the model with an architecture that is called Cross-Stitch units. Then these units are trained from end-to-end by combining the activations from multiple networks. Cross-Stitch units support a network to learn a suitable composition of shared and task-specific representations. The generalization across multiple tasks by the model dramatically improves its performance over the other methods with few training examples. This architecture diagram is shown in **Fig. 1.4.12**. Pooling and Fully-connected layers are placed after the Cross-Stitch units.

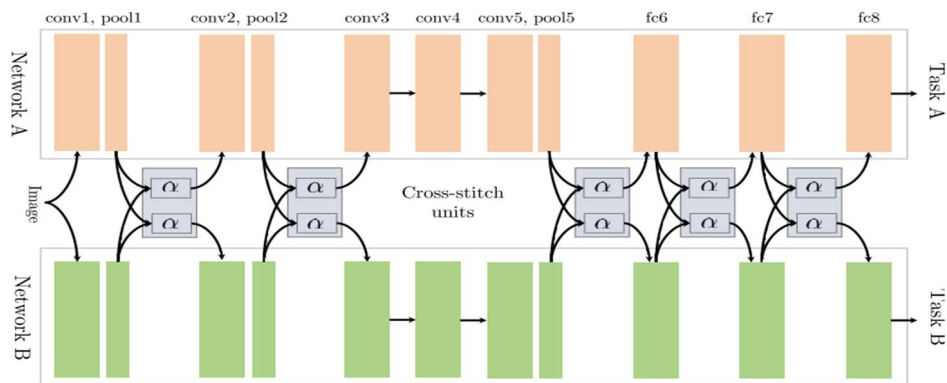

**Fig. 1.4.12**: Cross-Stitch Networks [22]

**1.4.4.2.4. Weighting losses with uncertainty**

Kendall et al. [23] adopt an alternative approach to understanding the structure of sharing by taking the possibility of each task into account. The weight of each task is then managed relative to a error function by calculating a multi-task cost function by improving the maximum likelihoodpossibility of tasks.They showed that the suggested model outperforms other trained models individually efficiently on their assigned task by learning multi-task weights. The proposed architecture can be seen in **Fig. 1.4.13**.

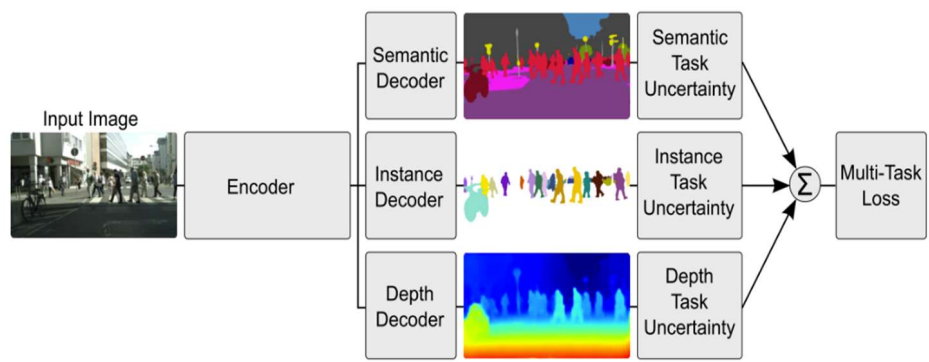

**Fig. 1.4.13**: Weighting losses with uncertainty [23]

## 1.4.4.2.5. Tensor Factorization for MTL - Sluice Networks

Some advances were made to generalize existing methods of Multi-tasking in Deep Learning. The method split the parameters of the model into task-specific and shared for each layer by applying tensor factorization. Yang and Hospedales [24] generalized some of the previously discussed matrix factorization approaches. Sluice networks are proposed in [25]. In this model, generalized DL-based MTL approaches such as block-sparse regularization approaches, hard parameter sharing, Cross-Stitch networks, and recent Natural Language Processing (NLP) methods create task hierarchy. The forwarded model diagram is shown in **Fig. 1.4.14**.

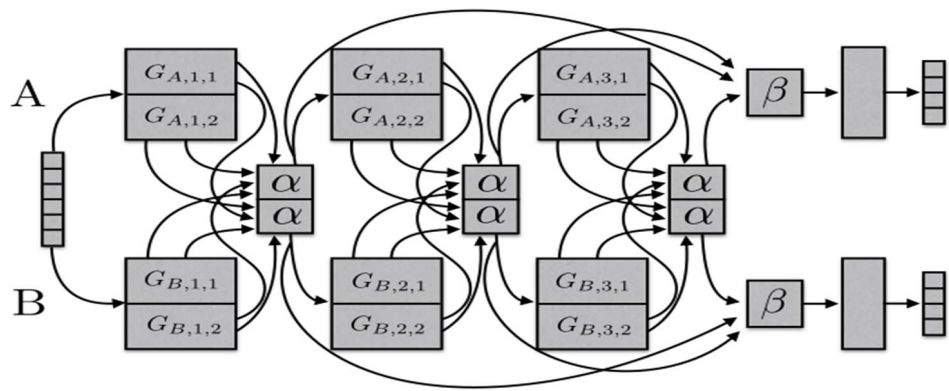

**Fig. 1.4.14**: Sluice Networks [25]

This proposed model allows it to identify the shareable layers and sub-spaces and to share them. This model architecture also determines which layers must have extracted the best description pattern of the supplied input sequences.

## 1.4.4.2.6. Joint Many-Task Model

Based on the previous findings, Hashimoto et al. [26] propose a hierarchical design comprising many NLP tasks, as seen in **Fig. 1.4.15**, as a joint MTL model. The approach used simple regularization terms to allow them to optimize all model weights. This helps in improving one task's loss without seriously affecting the performance of other tasks. This model can provide impressive results using parsing, relatedness, tagging, and entailment tasks.

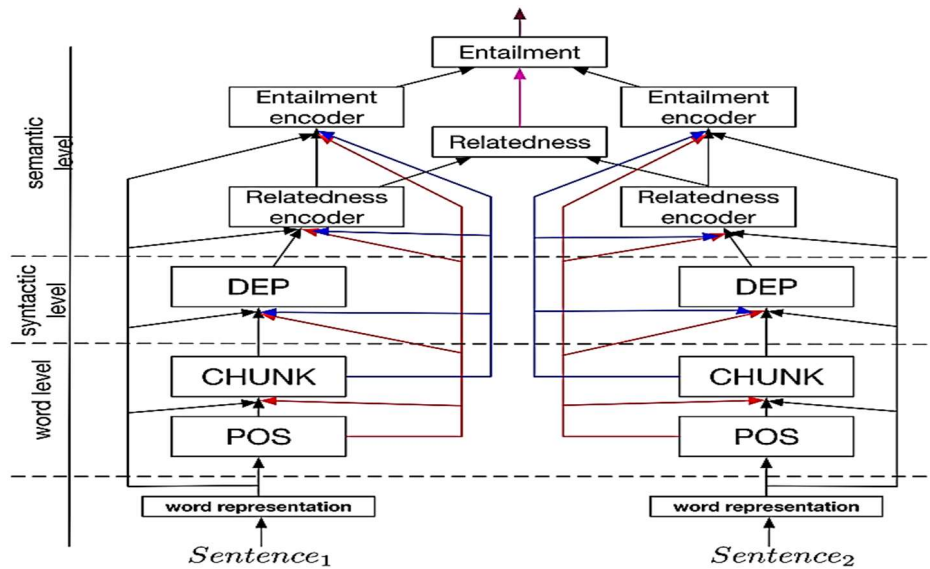
**Fig. 1.4.15:** Joint Many-Task Model [26]

### 1.4.4.3. Common Problems with Design and Implementation

#### 1.4.4.3.1. Combining Losses

The first challenge of designing the model is to define a single loss for the multi-task model. A single task may have a well-defined loss, but while designing the model for more than one task, selecting a single loss function might not be efficient. A poor selection of loss functions may lead the model to underperform. Over-fitting here will be less likely since the loss function will prevent the model from performing well even with a training dataset.

As a quick solution, summing the other losses can be tested for the dataset. However, there are specific problems with this approach. The sizes of the losses may be so diverse that one task might dominate the total loss while the other tasks have little chance of influencing the learning process in the common layers. A fast solution to this problem might be to replace the cumulative total of losses with a weighted sum, therefore leveling all losses to approximately the same magnitude. This method, however, includes another hyper-parameter that may need to be tweaked on occasion.

#### 1.4.4.3.2. Tuning Learning Rates

One of the essential hyper-parameters for tuning neural networks is the learning rate. This is a widespread convention. So, to improve the accuracy of a multi-task model, tuning can be tried to find a learning rate that will look excellent for a task. However, choosing a higher learning rate may kill the performance of activation layers for any undertaken tasks. However, using a low learning rate can also bring a prolonged convergence rate to the model.

For solving this problem, a different learning rate for each task-specific subnet can be used along with a single learning rate for the shared subnet.

### 1.4.4.3.3. Using Estimates as Features

Following the initial step of developing a deep learning model capable of predicting several tasks, the next job would be to apply an estimate for one task as a feature to another. In the forward pass, this is simple. Consider the following two tasks: A and B. Assume that the estimate for task A is provided to task B as a feature. However, we already have a label for A, so we are unlikely to propagate the gradients from task B to task A. This approach is helpful for deep learning models that need to multitask.

### 1.4.5 Applications

MTL has found applications in various fields. Following are some of the essential domains where multi-task learning can be applied.

### 1.4.5.1. Image Domain

Multi-tasking has been able to flourish in the image domain. Its benefits can be found in semantic scene parsing, pose estimation, dynamic MR image construction, diagnostic medical imaging, etc. The majority of the semantic scenes parsing models use supervised learning that depends on densely labeled data and does not promise good precision. The amount of data presently available for this field is also less, hindering performance measures of the prevailing deep learning models and other computer vision solutions. In the Semantic analyzing scene, the main challenge in the full textured RGB images is the highly variable content it contains in it.

Fourure et al. [27] introduced a gradient reversal strategy for domain adaptation. The target data from multiple datasets and various tasks is utilized using a particular loss function, and strong correlations among tasks are captured using an auto-context procedure. Yu and Lane [28] proposed a multi-tasking deep learning model which performs two image processing tasks: image classification and classification. The accuracy of the segmentation task doubled at pixel level while using the multi-tasking model instead of a single task image processing model. The accuracy of the classification task is also enhanced by 2% while using the multi-tasking model compared to a single task model. Moeskops [29] trained a multi-tasking CNN-based network to segment six tissues in brain MR images. The CNN learns to identify the anatomical structures of the brain, the imaging modality, and tissue classes. The model learns the diverse segmentation without the use of any task-specific training.

Multi-tasking Dynamic Contrast-Enhanced [30] imaging is used for the quantitative evaluation of neo-vascular architecture of the carotid artery walls. A strong correlation has been found between images which induced the images to be linearly dependent. Fully supervised deep neural networks typically require many pixel-level labels that are pretty costly to generate manually. Ke [31] put forward a deep learning multi-task model using a recursive approximation of tasks to reduce the manual cost of pixel-level labels. They divided the segmentation problem into a series of sub-problems defined recurrently based on improving performance in approximation. Coarse partial masks only labeled the majority of the images used in training. The model learns these partial masks statistics in the training phase. These regions are expanded towards the object's boundaries backed by the information learned in a data-driven manner.

Multi-tasking has also been used for real-time image enhancement [32]. A bilateral guided-up sampling approach has been proposed. The overall model is based on encoder-decoder architecture, and image processing shares all the encoding and decoding components. This multi-tasking model achieves higher performance compared to deep learning based on joint up-sampling methods. MTL has also found applications in local seismic image processing [33]. There are three basic image processing tasks related to seismic images- detecting a fault in structures, estimating the normal seismic vector to evaluate local orientation in structures, and removing noise by smoothing. These processing tasks are connected as they are based on analysis of the same seismic structural features. However, traditional techniques used for seismic image processing consider these tasks independent and face considerable challenges.

The idea of pooling information from multiple datasets using multi-tasking can be used in MNIST-like image datasets for image classification. The information sharing across hidden layers gives the potential of better learning and hence increased performance. The medical image processing domain suffers from collecting large datasets with clean annotations due to the limited availability of experts required for data labeling. Le et al. [34] proposed a deep learning based MTL model for picture classification and segmentation in mammography cancer detection, combining pixel-level segmentation with global-level classification annotations. The model is built on FNN, which allow for greater feature sharing and faster prediction. Multi-tasking is also used for CT image-based analysis for Covid-19 [35]. Instead of performing the classification and segmentation independently, they modeled an architecture composed of two decoders and an encoder for segmentation and reconstruction. AFNN is used for classification purposes.

### 1.4.5.2. Text Domain

DL has achieved many milestones in the domain of NLP, but the main drawback of the existing models is that they have to be trained from scratch, requiring large datasets and high computational costs. Present researches on NLP are mainly focused on transductive learning. A new model by Howard et al. [36] is based on inductive learning for text classification called Universal Language Model Fine-tuning for Text Classification. It has three stages: the first is used to build a pre-training language model using general domain data. The second stage is used for fine-tuning the language model for the target dataset. The last stage is the fine-tuning classification of the target data set.

Mrkvsic et al. [37] described a multi-tasking RNN model which shares hid-den layers among various tasks for training models involving multiple domain dialog state tracking. It learns from the most frequent and general dialogue features acquired across various domains. Collobert and Weston [38] proposed a multi-tasking neural network containing time-decay constraints to learn six NLP tasks. The tasks jointly include tagging part-of-speech, recognizing named entity, chunking, labeling of semantic role, language modeling, and related word identification. The model is trained by using unlabeled data. Wu and Huang [39] put forward a multi-tasking model for multi-domain sentiment classification that consists of a common sparse learner that can learn all the tasks and also tasks specific learners.

Multi-tasking is also used for sequence-to-sequence learning in the text domain. Luong et al. [40] used three multi-tasking settings for sequence-to-sequence based models; the first architecture involves one encoder, which is shared by all tasks and multiple decoders for specific tasks. The next one involves multiple encoders. Each task having its encoder and one decoder, which is shared among all tasks and the last one is a more general one that involves task sharing among multiple encoders and decoders.

Above mentioned applications of multi-tasking in the text domain indicates that multi-tasking has been able to give a satisfactory result in this domain. Transfer learning among tasks has been observed as a great advantage to the learning process. It requires less training data and less computational costs. However, this field is quite under-explored, and many new multi-tasking applications are possible in this domain. One can be improving language model pre-training, augmentation of language model with additional tasks, tasks where the amount of labeled data is limited.

### 1.4.5.3. Others

Apart from the text and image domain, multi-tasking is also used in bioinformatics, health informatics, speech processing, inverse dynamics problem for robotics, in stocks, climate prediction, etc. [41]. Multi-tasking is also used for modeling organisms [42]. Another multi-task model has been used to predict cross-platform si-RNA efficacy [43]. A brain-computer interface is constructed [44] based on multi-task learning without using any fixed calibration process for each subject. Xu et al. [45] used multi-task based approach for predicting location of sub-cellular protein. A multi-task based approach has been taken for information sharing about known disease genes [46] and prioritization of disease genes by learning from the shared information. Multi-task time series problems can be formulated for the prediction of Alzheimer's disease [47] where at each point of time for parameters organized in the matrix, a learner is associated. Li, et. al., [48] constructed as a multi-tasking classification problem for survival analysis by assuming that an event that occurs once will not take place once again, and MTFS is expanded for solving this problem.

Multi-tasking deep learning models have also been forwarded for speech synthesis [49]. A stacked deep neural network consisting of numerous neural networks out of which the preceding neural network is used to feed the result of the upper-most hidden layer to the upcoming layer. Each network has two inputs, the first one for the prime task and the other for additional tasks. The multi-task boosting method has been proposed for learning web search ranking [50] by sharing feature representation between various tasks. In [51] matrix rating in multiple related domains is done by a multiple domain collaborative filtering method. The MTRL method [52] is forwarded to consider the hierarchical structure and the sparsity in structure for maximum conversion in display advertising. A multi-task neural network is used for predicting a yearly return of stocks [53]. This model shares hidden layers for different prediction tasks. A multi-task model has also been proposed for multi-device localization [54] where learning is acquired from a low-rank transformation. It imposes similar model parameters for different tasks in the transformed space. Multi-tasking is used for solving the inverse dynamics problem in robotics [55]. Hence it can be concluded that multi-task learning has an excellent advantage in every aspect.

### 1.4.6. Evaluation of Multitask Models

For any multi-task model, the primary requirement is to learn whether the model can generalize on unknown data or memorize the data. We normally check our model performance on data other than the training data because the model can easily predict the value for any training data by learning it during the training time, which may lead to over-fitting. Therefore, we use a re-sampling method for the evaluation of the model. Few re-sampling methods to evaluate multi-tasking models are discussed below:

*a) Single hold-out random sub-sampling:* In hold-out, we primarily evaluate our model on different data. We divide the whole dataset into two parts: training data comprising about 70\% to 90\% of the dataset and test data comprising 10\% to 30\% of data. When both the training and test data are extensive, this is a reliable method for evaluation

*b) K-fold cross-validation:* The dataset is divided into k equal disjoint subsets in this technique. Among these k subsets, k-1 subsets are taken for training, and the rest is taken for testing purposes. The technique is performed k times, with each subset serving as testing data once. Then the average of the model's performance on these validation data is considered the model's overall performance.

*c) Leave one out cross-validation:* In this technique, one data unit is taken for validation, while the others are training data. The process is repeated for each data unit in the dataset. This technique incurs a substantial computational cost. It can be considered a special case of k-fold cross-validation when k=n, n being the number of data points in the dataset.

After considering a suitable method for re-sampling the data, the evaluation is done based on performance metrics. We choose a performance metric based on the multi-tasking task, i.e., classification, clustering, ranking, regression, etc.

Following are some of the powerful metrics used by the chapter we have explored:

*a) Accuracy:* It is a simple statistic that is used to assess performance. It is calculated as the total number of correct guesses to the total number of forecasts made.

*b) Precision:* The ratio of the total number of actual positive predictions to overall positive predictions.

*c) Recall:* It is calculated as the positive correctly predicted observations to the actual number of observations in the current class.

*d) F-measure:* F-measure is calculated as the harmonic mean of precision and recall. It is used when there is an uneven distribution of class like segmentation tasks.

*e) Confusion Matrix:* This metric is normally used for the evaluation of classification-based tasks. Here the number of predictions made for each class is shown in a matrix. The diagonal elements show the values which are correctly predicted (true positive).

*f) Root Mean Square Error (RMS):* It is calculated as the standard deviation of the prediction errors. In other words, it is the root of the Euclidian distance between the vector of anticipated values to the vector of values observed.

*g) The Area under the Curve (AUC):* It is a technique used by the binary classifier to measure performance by differentiating the true and false classes. A curve is drawn for a square of 1 unit, and the area contained by this curve is the model's accuracy.

Ke et al. [31] used the F1 score or dice coefficient to measure the performance of segmentation on the SEM database. Le et al. [34] used the Dice score to evaluate the performance of the segmentation task in cancer mammography, and they used the area under curve metric for measuring the performance of the classifier in predicting cancer. Wang et al. [47] used RMSE between predicted and actual cognitive scores to measure performance. Mrksic et al. [37] used the geometric mean of the classification accuracy to find the goal accuracy of the shared domains. Ranjan et al. [10] used mean average precision to measure the performance of face detection on AWS and PASCAL datasets, while normalized mean error is used for landmark localization. Das et al. [56] have found the accuracy for three classification tasks: gender, age, and ethnicity on UTK face dataset and BEFA challenge dataset.

### 1.4.7. Conclusion and Future Directions

Major Applications of Machine Learning such as Speech Recognition, Computer Vision, Drug Discovery, and Natural Language Processing have successfully benefited from the use of MTL. Doing MTL is more effective when compared to single-task learning when he/she has to optimize more than one loss function. Those cases will undoubtedly help to represent the problem in terms of MTL explicitly and gain from its models.

This chapter follows the advancement of diverse types of MTL methods and a brief description of each type. By using deep learning models, we usually intend to learn a better description of the traits/attributes of the input data to predict a definite value. Formally, we try to optimize for a singular function by training a model and fine-tuning the hyper-parameters up to the point where the performance increment is zero. In other words, it cannot be increased any further. By using MTL, it is also feasible to increase the performance even more by forcing the model to learn a more generalized representation of the tasks as it learns by updating its weights not just for one specific task but a group of tasks.

## Acknowledgment

We would like to acknowledge the authors Long et al., Lu et al., Misra et al., Kendall et al., and Ruder et al., for the architecture design diagram of their approach. This research was funded under grant number: ECR/2018/000204 by the Science and Engineering Research Board (SERB), Department of Science and Technology (DST) of the Government of India.